\documentclass[11pt]{article}

\usepackage[final]{acl}

\usepackage{times} 
\usepackage{latexsym}

\usepackage{amsmath}
\usepackage{amssymb}
\usepackage{array}
\usepackage{booktabs}
\usepackage{tabularx}

\usepackage{listings}

\newcolumntype{Y}{>{\raggedright\arraybackslash}X}

\usepackage[T1]{fontenc}

\usepackage[utf8]{inputenc}

\usepackage{microtype}

\usepackage{inconsolata}

\usepackage{graphicx}

\title{Which Mathematical Context Supports Target-Aligned Conjecture Generation? A Paired Ablation Study}

\author{
\textbf{Austin LaHue\textsuperscript{1}},
\textbf{Michael Burr\textsuperscript{1}},
\textbf{Luis David Garcia Puente\textsuperscript{2}}, \\
\textbf{Vinita Gangaram Jansari\textsuperscript{1}},
\textbf{Benjamin Nye\textsuperscript{2}},
\textbf{Carlos Toxtli Hernandez\textsuperscript{1}} \\
\textsuperscript{1}Clemson University,
\textsuperscript{2}Colorado College \\
\texttt{jlahue@clemson.edu}
}

\begin{document}
\maketitle
\begin{abstract}
Generating mathematical conjectures from natural-language context is challenging because theorem-relevant information is distributed across multiple forms of mathematical context, including definitions, objects, assumptions, and constructions. We study which of these contextual signals help a model recover the specific claim supported by the surrounding setup, using 61 main theorems from 12 papers across four mathematical domains. For each theorem, we redact the result and represent the remaining setup using six context components. A paired ablation design compares full context, six leave-one-out conditions, and six single-component conditions, yielding 793 generated conjectures. Context components were largely complementary: removing one component usually had little effect, whereas retaining only one substantially reduced expected-theorem alignment. Local Mathematical Objects provided the strongest standalone signal, while Structural Constraints and Conditions produced the largest alignment decrease when removed. These findings suggest that target-aligned conjecture generation depends on combining information that identifies relevant objects with information that constrains admissible claims.
\end{abstract}

\section{Introduction}

Mathematical conjectures are formed within a local discourse of definitions, objects, assumptions, examples, and established results. Automated conjecturing has traditionally searched structured symbolic, programmatic, or formally verified spaces \citep{raayoni-etal-2021-ramanujan,romera-paredes-etal-2024-funsearch,lin-etal-2024-atg}, while recent mathematical language systems show that supplied definitions, theorems, and repository context can support proof generation and formal reasoning \citep{welleck-etal-2022-naturalprover,hu-etal-2025-minictx}. Less is known about which parts of natural-language mathematical context help a model formulate the particular claim that a paper goes on to prove.

We study this question through \emph{target-aligned conjecture generation}. Given a redacted representation of the setup preceding a theorem, a model generates a formal conjecture without access to the theorem statement or proof. The task is narrower than open-ended mathematical discovery: success concerns whether the supplied context supports a mathematically substantive statement aligned with an existing hidden theorem. This controlled setting separates the ability to produce a plausible conjecture from the ability to recover the specific claim supported by the source context.

We conducted a paired ablation study over 61 main theorems from 12 papers spanning Commutative Algebra, Algebraic Geometry, Combinatorics, and Statistics Theory\footnote{Supplementary materials are available at \url{https://github.com/alahue/conjecture-ablation}.}. The redacted context is organized into six components and evaluated under full-context, leave-one-out, and single-component conditions. Results indicate that these components are largely complementary: removing one component generally had little effect, whereas retaining one alone substantially reduced alignment. Local Mathematical Objects supplied the strongest standalone signal, while removing Structural Constraints produced the largest alignment loss, suggesting distinct roles in identifying what a conjecture concerns and disambiguating what can be claimed. Together, these findings provide a component-level account of how theorem-local natural language supports target-aligned conjecture generation.

\section{Related Work}
\label{sec:related-work}

\subsection{Automated Conjecturing}

Prior work on automated conjecturing includes heuristic discovery over graph invariants, theory exploration, and search within restricted formula spaces \citep{fajtlowicz-1988-graffiti,larson-van-cleemput-2016-automated,colton-etal-2000-interestingness,mccasland-etal-2017-mathsaid,raayoni-etal-2021-ramanujan}. More recent systems combine learned guidance with executable search or formal verification \citep{romera-paredes-etal-2024-funsearch,lin-etal-2024-atg,dong-ma-2025-stp}. These approaches largely generate candidates in structured, programmatic, or formal spaces. We instead study conjecture generation from the mixed natural and symbolic language preceding theorems in research papers, asking which semantic components of that local discourse support target alignment.

\subsection{Mathematical Context}

Mathematical text interleaves prose, formulas, local notation, and document-level dependencies. Work on premise selection and identifier grounding shows that definitions, theorems, lemmas, propositions, formula structures, and co-references help identify the information relevant to a mathematical task \citep{ferreira-freitas-2020-premise,ferreira-freitas-2021-star,asakura-miyao-2024-needed}. Context-grounded proof systems use such local and repository-level information to support proof generation \citep{welleck-etal-2022-naturalprover,yang-etal-2023-leandojo,hu-etal-2025-minictx}, while autoformalization and semantic-alignment work connects informal mathematical language with formal statements \citep{wu-etal-2022-autoformalization,lu-etal-2025-formalign}. Our study differs by fixing a hidden theorem target and ablating semantically defined components of its preceding setup. Because long-context use and mathematical reasoning can vary with information position and surface form \citep{liu-etal-2024-lost,williamson-etal-2025-syntactic}, we interpret the observed effects as properties of the constructed prompts rather than of mathematical context in general.

\subsection{Ablation and Leakage}

Our design parallels rationale-faithfulness evaluation: removing evidence measures its conditional contribution, whereas retaining it alone measures sufficiency \citep{deyoung-etal-2020-eraser}. Accordingly, leave-one-out conditions estimate a component's contribution within rich context, and single-component conditions estimate standalone informativeness; the rankings may differ when information is complementary or redundant. Ablations can also create out-of-distribution inputs or preserve cues that directly reveal the target \citep{hase-etal-2020-leakage,hsia-etal-2024-goodharts}. We therefore avoid causal interpretations and separately audit redacted seeds and model-generated generalization axes for theorem-shaped leakage.

\subsection{Evaluation and Contamination}

Generated conjectures must be assessed both for correspondence to the source theorem and for mathematical quality. Rubric-guided LLM judges can approximate human evaluation \citep{liu-etal-2023-geval,zheng-etal-2023-judging}, but scores remain sensitive to judge choice, repeated runs, and irrelevant persuasive language \citep{haldar-hockenmaier-2025-rating,hwang-etal-2025-trick}. We report repeated-judge, cross-model, and expert agreement and treat scores as prompt- and judge-dependent estimates rather than verified correctness. Finally, reconstruction may reflect pretraining exposure. Contamination work distinguishes exposure from exploitation and advocates perturbation-based controls \citep{magar-schwartz-2022-data,qian-etal-2024-varbench,gulati-etal-2025-putnam}. Our publication-date holdout is exploratory, not proof of training-set inclusion or exclusion.

\section{Method}
\label{sec:method}

\subsection{Paper and Theorem Selection}

We sampled 12 arXiv papers from four mathematical subject domains,
with three papers selected from each:
\begin{itemize}\setlength{\itemsep}{0pt}\setlength{\parskip}{0pt}
    \item \texttt{math.AC} (Commutative Algebra)
    \citep{grimpen-etal-2026-relative,clarke-kosta-2024-distance,
    friedman-etal-2025-squared};
    \item \texttt{math.AG} (Algebraic Geometry)
    \citep{jeronimo-etal-2024-geometric,draisma-etal-2013-euclidean,
    salizzoni-etal-2025-nonlinear};
    \item \texttt{math.CO} (Combinatorics)
    \citep{sarkar-etal-2016-partial,le-romer-2023-equivariant,
    hollering-etal-2024-hyperplane};
    \item \texttt{math.ST} (Statistics Theory)
    \citep{brysiewicz-kim-2026-permutation,
    alexandr-etal-2022-decomposable,
    duarte-solus-2021-representation}.
\end{itemize}

Within each domain, papers were additionally required to have at least one Mathematics Subject Classification (MSC) code in one of the following families: 13Pxx (Computational aspects and applications of commutative rings), 14Qxx (Computational aspects in algebraic geometry), 68Wxx (Algorithms in computer science), or 62Rxx (Statistics on algebraic and topological structures). Among the selected papers classified within 62Rxx, all were assigned to 62R01 (Algebraic statistics).

For each paper, the theorem-selection stage of our pipeline classified all theorem-like environments and identified the paper’s \emph{main proved theorems}. A result qualified if it was presented as a central contribution of the paper, proved within the paper, and not merely a quoted result, supporting lemma or proposition, corollary, example, or background theorem. This process identified 61 theorem targets across the 12 papers. Individual papers contributed between 3 and 7 targets. The domain totals were 14 for \texttt{math.ST}, 17 for \texttt{math.AC}, 16 for \texttt{math.AG}, and 14 for \texttt{math.CO}.

Four of the selected papers, comprising 21 theorem targets, were posted after May 2025, a knowledge-cutoff date reported for DeepSeek-V4-Pro by third-party model documentation \citep{ng2026deepseekv4pro}. We treat these papers as a temporal-holdout cohort and compare them with the 8 earlier papers.

A human reviewer checked all 61 selected theorem targets against the source papers and confirmed that each was an eligible main proved theorem. We did not assess whether the pipeline identified every eligible theorem in the selected papers. One statement was incompletely extracted, while minor transcription issues included lost \LaTeX{} formatting in two papers and added definitions in two papers.

\subsection{Conjecture-Generation Pipeline}
\label{generation}

\begin{figure}[t]
    \centering
    \includegraphics[width=\columnwidth]{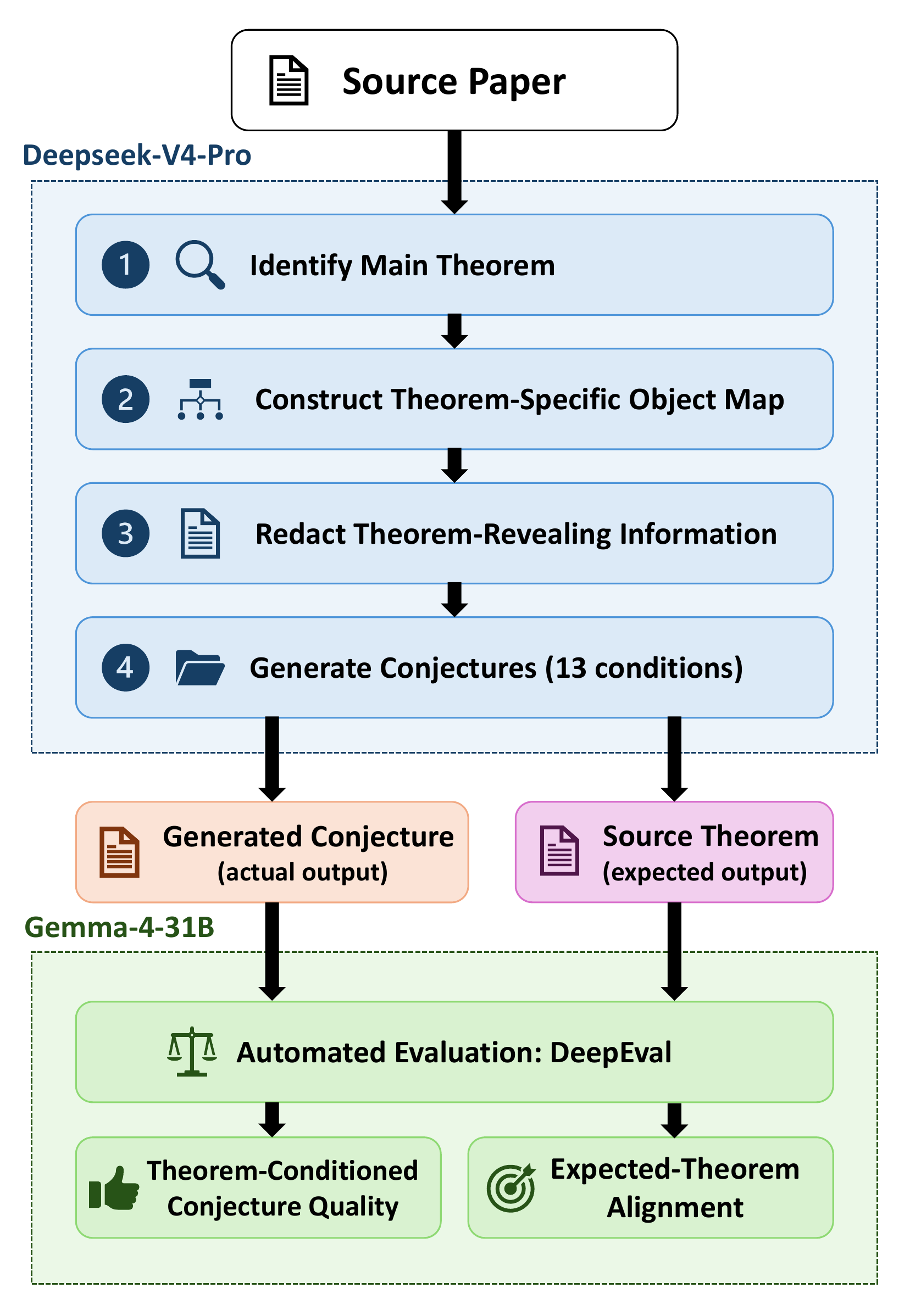}
    \caption{Overview of the conjecture-generation and evaluation pipeline.}
    \label{fig:pipeline}
\end{figure}

Figure~\ref{fig:pipeline} summarizes the end-to-end pipeline. We used Codex \citep{openai2025codex} as the orchestration harness, configured to call DeepSeek-V4-Pro \citep{deepseekai2026deepseekv4} through a vLLM-served endpoint \citep{kwon-etal-2023-vllm}. DeepSeek-V4-Pro performed theorem identification, object-map construction, redaction, and conjecture generation under each experimental condition. The generated conjecture and corresponding source theorem were then provided to DeepEval \citep{ip-vongthongsri-2026-deepeval}, with Gemma-4-31B \citep{gemmateam2026gemma4} serving as the judge, to produce theorem-conditioned conjecture-quality and expected-theorem-alignment scores.

For each theorem target, the generation pipeline performed the following
steps:

\begin{enumerate}
\item Construct a theorem-specific object map containing the source theorem, relevant definitions and notation, mathematical objects, assumptions, supporting results and constructions, and proof dependencies.
\item Convert the object map into a redacted seed by removing information that could reveal the target theorem. The redacted content included the theorem statement and conclusion, theorem identifiers, proof strategy, dependency chain, final classifications, and exact result-specific formulas or equivalences.
\item Organize the remaining setup information into six mathematical-context components.
\item Generate one conjecture from the selected components for each experimental condition.
\end{enumerate}

The conjecture-generation model was provided only with the redacted seed for the corresponding condition. The original paper, theorem map, unredacted object map, source theorem, proof strategy, and dependency chain were not included in the model input, and web search was disabled. It generated exactly one formal conjecture per theorem and condition. The substantive classification, redaction, and generation instructions are
summarized in Appendix~\ref{app:prompt-protocol}.

Table~\ref{tab:generation-example} shows a source theorem identified by the pipeline and the conjecture generated from its redacted mathematical-context components.

\begin{table}[b!]
\centering
\small
\setlength{\tabcolsep}{4pt}
\renewcommand{\arraystretch}{1.12}

\begin{tabularx}{\columnwidth}{@{}X@{}}
\toprule
\textbf{Source theorem} \\
\midrule

\begin{minipage}[t]{\linewidth}
Let \(U \subseteq V \subseteq R^d\) be submodules and
\(\prec\) a fixed monomial order on \(R^d\). Then \(V\) has a unique
reduced Gröbner basis relative to \(U\).

\vspace{4pt}
\end{minipage}
\\

\midrule
\textbf{Generated conjecture} \\
\midrule

\begin{minipage}[t]{\linewidth}
Let \(\Bbbk\) be a field, let
\(R = \Bbbk[X_1, \dotsc, X_n]\), and let
\(U \subseteq V \subseteq R^d\) be nested submodules of the free module
\(R^d\). Fix a monomial order
\(\prec\) on \(\operatorname{Mon}(R^d)\). Then there exists a reduced
relative Gröbner basis \(H\) of \(V\) with respect to \(U\).

Moreover, this reduced relative Gröbner basis is unique: if \(H\) and
\(H'\) are two reduced relative Gröbner bases of \(V\) relative to
\(U\), with respect to the same monomial order \(\prec\), then
\[
H = H'.
\]

\vspace{-8pt}
\end{minipage}
\\

\bottomrule
\end{tabularx}
\caption{Example of a source theorem \citep{grimpen-etal-2026-relative} and the corresponding generated conjecture.}
\label{tab:generation-example}
\end{table}

\subsection{Mathematical-Context Components}

Each redacted seed represented the theorem's preceding context using six components:

\begin{enumerate}
\item \textbf{Ambient Setting}: the broad mathematical setting and background assumptions, without identifying the target theorem;
\item \textbf{Core Definitions and Notation}: definitions and notation required to understand the local problem;
\item \textbf{Local Mathematical Objects}: structures, quantities, maps,
operators, relations, functions, families, or other mathematical entities
that define the local problem setting;
\item \textbf{Structural Constraints and Conditions}: hypotheses, restrictions, and validity conditions defining the local problem setting;
\item \textbf{Auxiliary Objects and Constructions}: supporting examples,
intermediate objects, derived constructions, or families used to study,
illustrate, or relate the local mathematical objects;
\item \textbf{Potential Generalization Axes}: model-generated directions for forming conjectures, such as varying parameters, weakening assumptions, changing the ambient object, or examining exceptional cases.
\end{enumerate}

The first five components were summarized from the theorem-specific source material after redaction. The Potential Generalization Axes were synthesized by DeepSeek-V4-Pro from the redacted mathematical setup.

All 61 redacted theorem seeds underwent a leakage audit (Appendix~\ref{sec:leakage-audit}). Fifty-eight seeds (95.1\%) were flagged for potential inferential leakage and reviewed, but all were retained after mitigation; no audit summary reported an explicit hidden theorem statement or final conclusion remaining in a seed.

Because Potential Generalization Axes were synthesized by the same model used for conjecture generation, we audited this component separately for theorem-shaped information (Appendix~\ref{sec:generalization-axis-audit}). Of the 61 seeds, 12 (19.7\%) were classified as direct theorem-shaped, 24 (39.3\%) as strongly directional, and 25 (41.0\%) as low or contextual. Accordingly, results involving this component may reflect theorem-directed cues in addition to genuine support for conjecture generation and are therefore interpreted cautiously.

\subsection{Ablation Design}

For each of the 61 theorem targets, we generated conjectures under 13 context conditions:

\begin{itemize}
\item one \textbf{full-context} condition containing all six components;
\item six \textbf{leave-one-out} conditions, each excluding one component while retaining the other five;
\item six \textbf{single-component} conditions, each retaining one component while excluding the other five.
\end{itemize}

This yielded $61 \times 13 = 793$ generated conjectures. The design is paired: all 13 conditions for a theorem were derived from the same theorem-specific object map and evaluated against the same source theorem. Leave-one-out conditions estimate a component's contribution in the presence of otherwise rich context, whereas single-component conditions estimate its standalone informativeness.

Because redaction cannot prevent recovery of a theorem from model pretraining, we conducted an exploratory temporal-holdout analysis. We compared full-context performance and the mean effects of the leave-one-out and single-component conditions between the four post-cutoff papers and the eight earlier papers. Cohort comparisons used paper-level summaries so that each paper received equal weight. Because DeepSeek has not published an official knowledge cutoff, publication date does not establish actual training-set inclusion, and the cohorts are not balanced by domain, we treat this analysis as exploratory.

\subsection{Automated Conjecture Evaluation}

We evaluated conjectures using DeepEval with Gemma-4-31B as the primary judge and a temperature of zero, following exploratory comparisons among candidate judges. DeepEval received the generated statement as the \emph{actual output} and the original source theorem as the \emph{expected output}. Scores were assigned on a scale from \(0\) to \(1\) in increments of \(0.1\), with higher values indicating stronger performance.

We used two metrics:

\begin{itemize}
\item \textbf{Expected-Theorem Alignment}: implemented using DeepEval's documented correctness criterion: ``Determine whether the actual output is factually correct based on the expected output.'' \citep{confidentai-deepeval-metrics} Because this measure compares a generated conjecture with a target theorem rather than independently verifying the conjecture, we interpret it as LLM-estimated theorem-alignment rather than verified mathematical correctness.
\item \textbf{Theorem-Conditioned Conjecture Quality}: a custom, target-aware rubric assessing whether the output contains a clear and substantive mathematical claim, its simple-case plausibility, precision, potential usefulness, and relevance to the source theorem. Unlike expected-theorem alignment, this metric can reward a meaningful specialization, close variant, or plausible generalization even when it does not accurately recover the target result. Because the expected theorem is available to the judge, this metric does not measure source-independent novelty or intrinsic conjecture quality. The complete rubric is provided in Appendix~\ref{sec:appendix}.

\end{itemize}

Both metrics are conditioned on the expected theorem, but operationalize
different outcomes. Expected-theorem alignment is reconstruction-oriented,
whereas theorem-conditioned conjecture quality permits mathematically
meaningful departures from the target theorem.

\subsection{Expert Validation}

We evaluated agreement between the automated theorem-conditioned quality judge and an expert mathematician using an independent validation set of 15 conjectures. These conjectures were generated separately using o3-mini \citep{openai2025o3mini} and were not sampled from the ablation-study outputs or assigned to experimental conditions. The expert applied the same theorem-conditioned quality rubric used by the automated judge and was blinded to the LLM-generated scores.

Expert and Gemma theorem-conditioned quality scores were moderately positively correlated, \(r(13)=.60\), \(p=.017\), 95\% CI \([.13,.85]\). Absolute agreement was also moderate using a two-way mixed-effects, absolute-agreement, single-measure intraclass correlation, \(\operatorname{ICC}(A,1)=.54\), with a bootstrap 95\% CI of \([.09,.77]\). The mean absolute error was \(0.27\), and Gemma assigned scores that were, on average, \(0.16\) points higher than the expert. We therefore interpret this validation as preliminary evidence of evaluator agreement rather than definitive validation, given the single expert, small sample, and use of an independently generated validation set.

The expert did not evaluate expected-theorem alignment because doing so would require substantial proof development or counterexample search. The Expert theorem-conditioned quality scores were positively associated with the automated expected-theorem-alignment scores, \(r(13)=.58\), \(p=.022\), 95\% CI \([.10,.84]\); however, because these variables measure different constructs, we treat this as an exploratory association rather than direct validation of the alignment metric.

\subsection{Judge Reliability and Cross-Model Agreement}

To assess evaluator reliability, we rescored the full-context and six leave-one-out outputs from four papers, with one paper selected from each domain. We first repeated the evaluation using Gemma-4-31B with the same prompts and settings. We then evaluated the same outputs using GPT-5-mini \citep{openai2025gpt5} as an alternative judge. For each metric, we assessed score association using Pearson's \(r\) and absolute agreement using \(\operatorname{ICC}(A,1)\).

We also piloted GLM-5.1-FP8 \citep{zai2026glm51fp8,glm5team2026glm5} as an expected-theorem-alignment judge. Its lower agreement with Gemma appeared to reflect a stricter treatment of claims absent from the expected theorem. We therefore report it as an exploratory diagnostic of judge and prompt sensitivity (Appendix~\ref{sec:glm-pilot}).

\subsection{Statistical Analysis}

Because each theorem target was evaluated under every context condition, analyses were paired at the theorem level. For leave-one-out conditions, we report mean differences from the full-context baseline, paired effect sizes (\(d_z\)), and 95\% confidence intervals for the paired differences. Overall differences among the full-context and six leave-one-out conditions were assessed for each evaluation metric using Friedman tests, followed by Wilcoxon signed-rank comparisons with the full-context baseline and Holm correction across the six comparisons.

Because theorem targets from the same paper share paper-level context, we treated them as clustered rather than independent observations. To prevent papers with more targets from receiving greater weight, we repeated the primary analyses using paper-level mean scores, with each paper weighted equally. Domain comparisons were computed from these summaries and treated as exploratory because each domain contained only three papers. For each metric, differences in component-removal effects across domains were assessed using Kruskal--Wallis tests with Holm correction across the six components. Single-component conditions were analyzed descriptively to estimate standalone informativeness.

\section{Results}
\subsection{Context Components Are Largely Complementary}

\begin{figure*}[t]
    \centering
    \includegraphics[width=\textwidth]{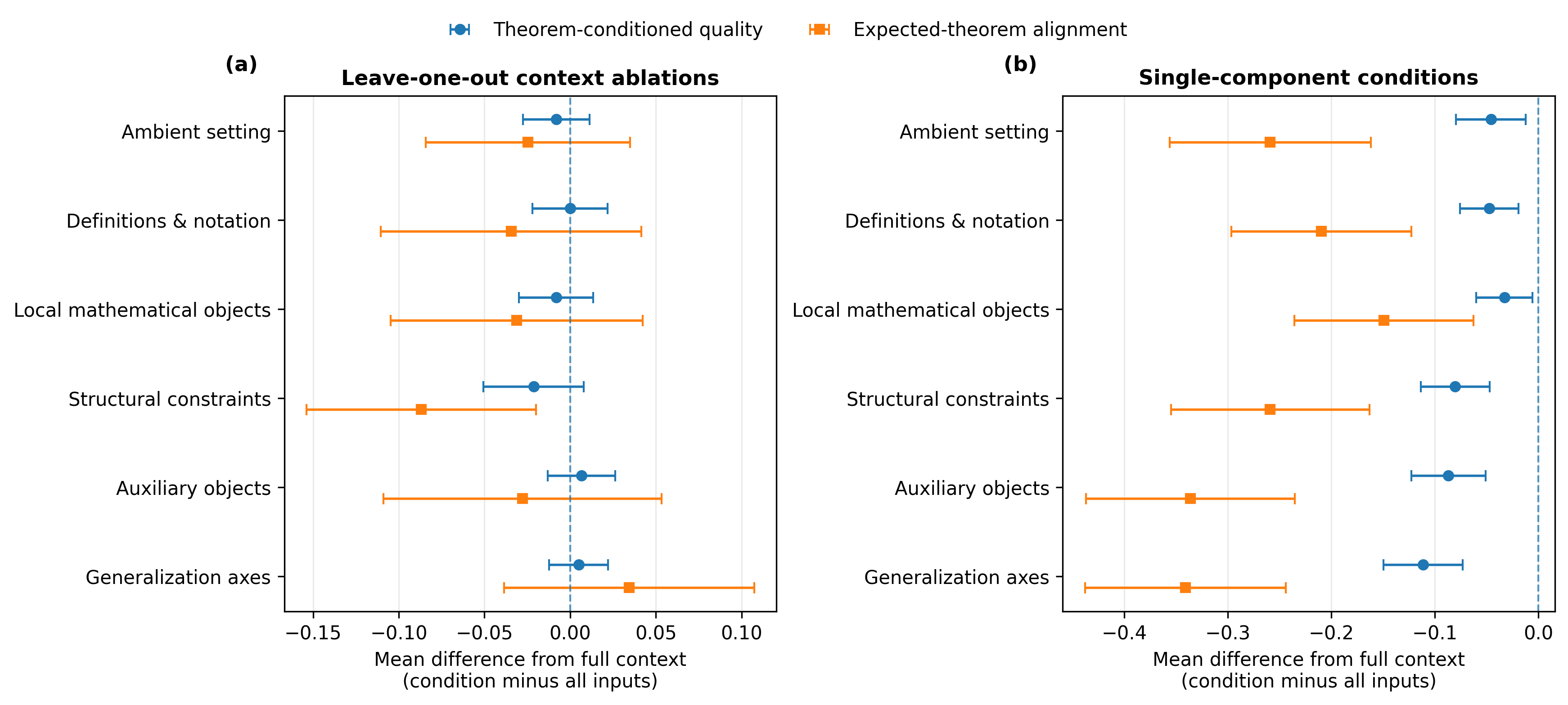}
    \caption{Paired theorem-level differences from the full-context condition
    with 95\% confidence intervals. Panel (a) removes one context component,
    whereas panel (b) retains only the named component. Negative values indicate
    lower scores than under full context.}
    \label{fig:overall-ablation}
\end{figure*}

Under the full-context condition, mean theorem-conditioned conjecture quality was \(0.931\), and mean expected-theorem alignment was \(0.607\). Figure~\ref{fig:overall-ablation} reports paired theorem-level differences from these baselines. Complete condition-level results are reported in Appendix~\ref{sec:complete-ablation-results}.

Leave-one-out ablations produced only small changes in either metric. The omnibus tests were not significant for theorem-conditioned quality, \(\chi^2(6)=5.28\), \(p=.509\), or expected-theorem alignment, \(\chi^2(6)=9.75\), \(p=.135\). Removing Structural Constraints and Conditions produced the largest alignment decrease, \(\Delta=-0.087\), 95\% CI \([-0.154,-0.020]\), but the comparison did not remain significant after Holm correction, \(p_{\mathrm{Holm}}=.055\). No other leave-one-out effect was clearly distinguishable from zero.

Equal-weight paper-level analyses yielded the same qualitative pattern. The omnibus tests remained nonsignificant, and removing Structural Constraints again produced the largest alignment decrease, \(\Delta=-0.085\), \(p_{\mathrm{Holm}}=.105\).

The single-component conditions revealed a distinction between standalone and complementary information. Local Mathematical Objects provided the strongest standalone signal, with an alignment mean of \(0.457\). Structural Constraints were less informative alone, with an alignment mean of \(0.348\), despite producing the largest decrease when removed from full context. Thus, Local Mathematical Objects supplied the most directly usable information, whereas Structural Constraints contributed by disambiguating otherwise rich context.

\subsection{Component Importance Varies by Domain}

\begin{table}[!b]
\centering
\small
\setlength{\tabcolsep}{3pt}

\begin{tabularx}{\columnwidth}{@{}l r X r X@{}}
\toprule
\textbf{Domain} &
\textbf{Full} &
\textbf{Largest loss} &
\(\boldsymbol{\Delta}\) &
\textbf{Best standalone} \\
\midrule
\texttt{math.AC}
& .611
& Constraints
& -.162
& Local objects \\

\texttt{math.AG}
& .571
& Constraints
& -.121
& Definitions \\

\texttt{math.CO}
& .644
& Definitions
& -.086
& Constraints \\

\texttt{math.ST}
& .664
& Local objects
& -.096
& Definitions \\
\bottomrule
\end{tabularx}
\caption{Exploratory domain-level expected-theorem alignment.
Full is the full-context baseline. Each domain contains three papers.}
\label{tab:domain-alignment-summary}
\end{table}

Domain-level analyses were exploratory because each domain contained only three papers. Full-context performance was broadly similar across domains, but the component associated with the largest alignment loss varied (Table~\ref{tab:domain-alignment-summary}). Structural Constraints were most consequential in the two algebraic domains, whereas Definitions and Notation were most consequential in \texttt{math.CO}, and Local Mathematical Objects were most consequential in \texttt{math.ST}. The complete domain-level results are reported in Appendix~\ref{sec:complete-domain-results}.

Potential Generalization Axes showed the clearest qualitative domain difference. Removing them reduced alignment in \texttt{math.AC} and \texttt{math.AG}, but increased alignment in \texttt{math.CO} and \texttt{math.ST}. The removal effect did not differ significantly across domains after Holm correction, \(p_{\mathrm{Holm}}=.553\).

\subsection{Held-Out Papers Show Suggestive Sensitivity to Context Reduction}
Full-context performance was similar for the post-cutoff and earlier-paper cohorts: expected-theorem alignment was \(0.630\) and \(0.619\), respectively, while theorem-conditioned quality was \(0.929\) and \(0.942\). However, the post-cutoff papers were more sensitive to incomplete context. Their mean leave-one-out alignment effect was \(-0.083\), compared with approximately zero for the earlier papers, and their mean single-component effect was \(-0.346\), compared with \(-0.218\). Theorem-conditioned quality effects were nearly identical between cohorts. The paper-level cohort differences were not significant for either the leave-one-out (\(p=.214\)) or single-component (\(p=.087\)) analyses. These descriptive patterns are consistent with greater reliance on the supplied context for temporally held-out papers, but do not establish whether any individual paper was present in the model's training data.

\subsection{Judge Agreement Varies by Metric}
Repeated Gemma evaluation showed high reliability for both metrics (Table~\ref{tab:judge-reliability}). Cross-model agreement with GPT-5-mini was lower: alignment scores showed moderate absolute agreement, whereas absolute quality scores were sensitive to judge choice.

\begin{table}[b!]
\centering
\small
\setlength{\tabcolsep}{5pt}
\begin{tabular}{@{}llcc@{}}
\toprule
\textbf{Comparison} & \textbf{Metric} & \(r\) & \textbf{ICC} \\
\midrule
Gemma repeat
    & Alignment & .921 & .921 \\
    & Quality   & .840 & .841 \\
\addlinespace
GPT-5-mini
    & Alignment & .784 & .731 \\
    & Quality   & .568 & .212 \\
\bottomrule
\end{tabular}
\caption{Repeated-judge reliability and cross-model agreement.}
\label{tab:judge-reliability}
\end{table}

\section{Discussion}

This study examined which forms of the theorem-local mathematical context support conjectures aligned with a hidden target theorem. Across 61 theorem targets and 13 paired conditions, the results indicate that context components are largely complementary rather than individually indispensable. Removing one component from an otherwise complete context generally produced only small changes, whereas retaining a single component substantially reduced expected-theorem alignment. Thus, the small leave-one-out effects do not imply that context is unimportant. Instead, multiple components appear to provide overlapping information and partially compensate for one another.

The contrast between the two evaluation metrics further clarifies this result. Full context produced high theorem-conditioned conjecture quality but only moderate expected-theorem alignment. Even under single-component conditions, the model often generated plausible and relevant mathematical statements despite failing to recover the particular result proved in the source paper. Limited context may therefore be sufficient for producing a mathematically substantive continuation, while richer context is needed for target-aligned generation. These are related but distinct capabilities.

\subsection{Objects Propose; Constraints Disambiguate}

The clearest functional distinction emerged between Local Mathematical Objects and Structural Constraints and Conditions. Local Mathematical Objects provided the strongest standalone alignment signal, suggesting that theorem-relevant objects, maps, quantities, and families help define the broad mathematical subject and were often sufficient for the model to produce a related claim.

Structural Constraints played a different role. Although less informative in isolation, their removal from the full context produced the largest decrease in expected-theorem alignment. This pattern suggests that constraints are primarily disambiguating: mathematical objects define a space of possible claims, while hypotheses, restrictions, and validity conditions help determine which relation among those objects is intended. Objects may therefore support candidate formation, whereas constraints help select among otherwise plausible candidates.

This distinction also shows why component importance cannot be represented by a single ranking. Standalone performance reflects how directly a component supports generation by itself, while leave-one-out effects reflect its contribution when combined with the remaining context. For target-aligned conjecture generation, direct informativeness and conditional contribution appear to be separate properties.

The exploratory domain results suggest that these functions may vary across mathematical fields. Structural Constraints produced the largest alignment losses in Commutative Algebra and Algebraic Geometry, while Definitions and Notation were most consequential in Combinatorics and Local Mathematical Objects in Statistics Theory. Because each domain contained only three papers, these patterns are not conclusive, but they motivate domain-adaptive approaches to selecting and organizing mathematical context.

The Potential Generalization Axes showed less consistent effects. Their removal reduced alignment in the algebraic domains but increased it in Combinatorics and Statistics Theory, and they were weak as a standalone component. Generalization directions may sometimes focus generation toward the target, but they may also broaden the candidate space. Because these axes were model-generated rather than directly extracted from the source, they are better treated as a conjecturing scaffold than as equivalent to the five source-derived components.

\subsection{Context Dependence and Evaluation}

Earlier and post-cutoff papers achieved similar full-context performance, but the post-cutoff cohort showed larger alignment losses when context was reduced. This pattern is consistent with the model relying more heavily on the supplied mathematical setup when the target material is less familiar from training. It does not establish whether any paper was included in the training data, and the cohort differences were not statistically conclusive. The temporal analysis therefore provides evidence about context sensitivity rather than contamination itself. Controlled perturbations, such as systematic symbol renaming or paraphrasing, would provide stronger tests of whether generation depends on supplied mathematical meaning, familiar lexical patterns, or prior exposure.

Evaluator agreement also differed by metric. Expected-theorem alignment showed stronger repeated and cross-model agreement than theorem-conditioned conjecture quality. Alignment asks a constrained question about correspondence to a supplied theorem, whereas quality is a broader judgment of clarity, simple-case plausibility, precision, potential usefulness, and relevance to the source theorem. Expected-theorem alignment therefore provides a more stable basis for comparing ablation conditions, while theorem-conditioned quality indicates whether low-alignment outputs remain mathematically substantive.

Neither metric independently establishes mathematical truth or novelty. Future evaluation could score the existing quality dimensions separately rather than combining them into one holistic judgment. It could also classify each output as equivalent to the target theorem, a specialization, a generalization, or a different but related result. Proof attempts, counterexample search, and source-blind expert review would further distinguish target recovery from mathematical validity.

Overall, the findings suggest that conjecture-generation systems may benefit from organizing context according to its mathematical role. Local Mathematical Objects provide an efficient basis for identifying what to conjecture about, while Structural Constraints, definitions, and ambient information narrow the admissible claims. Because component effects varied across domains, adaptive context assembly may be more effective than a fixed minimal prompt. Target alignment appears to emerge from the interaction between context that identifies relevant objects and context that constrains admissible claims.

\section{Conclusion}

We presented a paired ablation study of theorem-local context for target-aligned conjecture generation. Across 61 theorem targets, Local Mathematical Objects were most informative in isolation, while Structural Constraints contributed most within richer context. Future work should test whether these patterns generalize across additional models, domains, context decompositions, and source-blind evaluations of validity and novelty.

\section*{Limitations}

Both evaluation metrics are conditioned on the expected theorem. The theorem-conditioned quality score therefore does not measure source-independent novelty, usefulness, or mathematical validity, while expected-theorem alignment remains sensitive to judge and prompt interpretation. This sensitivity was reflected in the stricter behavior of GLM-5.1-FP8 relative to Gemma-4-31B and GPT-5-mini. Expert validation was also limited to one mathematician and 15 conjectures generated separately by o3-mini, so agreement may not generalize to the ablation-study outputs. The expert did not independently validate expected-theorem alignment, as doing so would require proof development or counterexample search. Accordingly, the reported scores should be treated as target-conditioned estimates rather than definitive assessments of conjecture quality or correctness.

The ablations do not isolate the causal importance of each component. Removing or retaining components changes prompt length, coherence, redundancy, and resemblance to natural mathematical exposition. Single-component prompts may be especially unnatural, while leave-one-out prompts may allow overlapping information to compensate for the removed component. The study is also limited to 61 theorem targets from 12 papers, four computationally oriented domains, one generation model, and one conjecture per theorem--condition pair, so within-condition generation variability was not estimated. Domain comparisons are therefore exploratory, and the results may depend on the selected papers, prompts, and component-construction procedure. The LLM-based selection pipeline was validated only for the targets it selected; because recall was not assessed, it may have omitted other eligible main theorems.

The temporal-holdout analysis relies on a third-party knowledge-cutoff estimate because DeepSeek has not disclosed an official cutoff for DeepSeek-V4-Pro. The publication date cannot establish whether a paper or related material appeared in training, and the temporal cohorts also differ in domain composition. These results should therefore be interpreted as evidence of differing context sensitivity rather than direct evidence of training-data contamination.

Potential Generalization Axes present an additional leakage risk because they were synthesized from theorem-specific material by the same model used for generation. The audit classified 19.7\% of seeds as directly theorem-shaped and 39.3\% as strongly directional. Effects involving this component may therefore reflect theorem-directed cues as well as support for generalization. More broadly, the six components represent one operationalization of mathematical context; other reasonable decompositions may produce different results.

\section*{Acknowledgments}
This work was supported by the National Science Foundation under Award No. 2432704. This research used in part resources on the Palmetto 2 cluster at Clemson University under National Science Foundation awards MRI 1228312, II NEW 1405767, MRI 1725573, and MRI 2018069. The views expressed in this article do not necessarily represent the views of NSF or the United States government.

The authors used ChatGPT to assist with refining the language and organization of the manuscript and to support discussion of statistical analyses and presentation of results. All generated suggestions were reviewed, verified, and revised by the authors, who take full responsibility for the content of the paper. Generative AI models used as part of the experimental methodology are described separately in Section~\ref{generation}.


\bibliography{references}

\newpage
\appendix

\section{Leakage Audit}
\label{sec:leakage-audit}

All 61 redacted theorem seeds across the 12 paper sets underwent a post-redaction leakage audit. Of these, 58 seeds (95.1\%) were flagged for at least one ambiguity or potential leakage concern. A flag indicated a residual inferential risk requiring review or mitigation rather than a failed redaction. After mitigation, all 61 seeds were retained. The audit achieved 100\% processing yield, with no skipped files, unmatched outputs, or input--output identifier mismatches, and no audit summary reported an explicit hidden theorem statement or final conclusion remaining in a seed.

The audit recorded 62 concern groups. The most common involved retained proof, algorithm, or construction architecture (40.3\%), followed by hints about equivalence, containment, equality, or characterization (21.0\%), quantitative formulas or thresholds (14.5\%), and structural or definitional setup that narrowed the likely result (12.9\%). Mitigation included removing explicit conclusions and quantitative answers, weakening proof dependencies, reframing revealing relationships as open questions or background, and separating definitions from their consequences. Thus, the remaining leakage risk is primarily inferential and contextual rather than explicit.

\section{Potential Generalization Axes Audit}
\label{sec:generalization-axis-audit}

Because Potential Generalization Axes were synthesized by the same model used
for conjecture generation, we conducted a separate content-flow audit to
assess whether this component could reintroduce information about the hidden
theorem. Each seed's axes were compared with the corresponding exact theorem
statement and assigned one seed-level severity:

\begin{itemize}
    \item \textbf{Direct theorem-shaped}: an axis names substantially the same
    objects, conditions, and conclusion or formula pattern, such that converting
    the question into a declarative statement approximately reconstructs the
    theorem.
    
    \item \textbf{Strong directional}: the axes reveal the theorem's type,
    architecture, relationship, or expected output, but do not reproduce the
    complete statement.
    
    \item \textbf{Low/contextual}: the axes vary assumptions, parameters,
    methods, or ambient settings without identifying the target conclusion.
\end{itemize}

The 61 seeds contained 450 generalization-axis bullets, with a mean of 7.38
bullets per seed. Table~\ref{tab:generalization-axis-audit} reports the
seed-level classifications.

\begin{table}[t]
\centering
\small
\begin{tabular}{lrr}
\toprule
\textbf{Risk level} & \textbf{Seeds} & \textbf{Share} \\
\midrule
Direct theorem-shaped & 12 & 19.7\% \\
Strong directional    & 24 & 39.3\% \\
Low/contextual        & 25 & 41.0\% \\
\midrule
Total                 & 61 & 100.0\% \\
\bottomrule
\end{tabular}
\caption{Audit of Potential Generalization Axes for theorem-shaped information.}
\label{tab:generalization-axis-audit}
\end{table}

Thus, 36 seeds (59.0\%) contained at least strong directional information,
including 12 whose axes approximately encoded the shape of the source theorem.
This audit characterizes information flow from the source theorem into the
constructed component; it does not measure how frequently a generation model
would actually reconstruct the theorem from that information.

\section{Exploratory GLM-5.1-FP8 Judge Analysis}
\label{sec:glm-pilot}

We piloted GLM-5.1-FP8 as an additional expected-theorem-alignment judge on the subset used for the repeated-Gemma and GPT-5-mini analyses. Its scores showed weaker agreement with Gemma, with \(r=.649\) and \(\operatorname{ICC}(A,1)=.307\).

An inspection of cases with high Gemma-4-31B and GPT-5-mini scores but low GLM-5.1-FP8 scores suggested that GLM adopted a stricter interpretation of the evaluation prompt. Its rationales frequently noted that the generated output introduced claims ``not present in or supported by the Expected Output,'' effectively treating the expected theorem as the evidentiary basis for every claim. Gemma-4-31B and GPT-5-mini were more permissive toward additional claims when the output otherwise corresponded closely to the target theorem.

We therefore treat the GLM-5.1-FP8 run as an exploratory diagnostic of judge and prompt sensitivity rather than part of the primary cross-model agreement analysis. This post hoc treatment was not based on a prespecified agreement threshold.

\section{Theorem-Conditioned Conjecture Quality Prompt}
\label{sec:appendix}
\begin{itemize}
    \item "Evaluate the actual output as a generated mathematical conjecture. The goal is to assign a score based on the quality, relevance, and mathematical usefulness of the conjecture."
    \item "Use the input as the context that was provided to the conjecture-generating model. Use the expected output as a theorem from a mathematics paper whose preceding text was used to generate the conjecture. Use the actual output as the generated conjecture being evaluated."
    \item "First, compare the actual output to the expected output. Determine whether the actual output is essentially the same as the expected theorem, a close variant, a related generalization or specialization, loosely related, or unrelated."
    \item "Second, remove fluff. Ignore generic motivational language, vague claims of importance, broad exposition, future-work language, and text that does not assert a mathematical claim. Determine whether a clear mathematical statement remains."
    \item "Assess whether anything mathematically meaningful remains after removing fluff. If the remaining content does not contain a clear mathematical statement, assign a low score. If it appears to contain a potentially new mathematical statement, continue evaluating it using the knownness, simple-case plausibility, and relevance criteria below."
    \item  "Assess simple-case plausibility. Penalize statements that are internally inconsistent, obviously false, missing necessary hypotheses, overgeneralized in a way that would likely fail immediately, or likely to fail on the first interesting example. Do not perform extensive computation; use basic mathematical reasoning only."
    \item "Assess knownness and novelty. A known result is less valuable as a new conjecture than a plausible new statement. However, do not overly penalize close recovery of the expected theorem: similarity to the expected theorem is evidence that the output is mathematically meaningful, interesting, and contextually relevant."
    \item "Assess relevance to the expected theorem. Outputs close to the expected theorem should generally receive higher scores. Outputs very different from the expected theorem should often receive C-level scores if they contain substantive mathematical content but are difficult to evaluate confidently."
    \item "Assign a continuous score from 0 to 1 using this scale: 1.0 means A-level: clear, substantive, plausible, highly relevant, and mathematically interesting; it closely recovers, usefully refines, or interestingly extends the expected theorem. 0.75 means B-level: good mathematical conjecture; mostly clear, plausible, and relevant, but less precise, less novel, or less strong than A. 0.5 means C-level: contains potentially interesting mathematical content, but its relevance, novelty, precision, or plausibility is uncertain; this includes many outputs that are quite different from the expected theorem but not clearly bad. 0.25 means D-level: weak conjecture; little substance remains after removing fluff, or the statement is vague, trivial, mostly known, poorly connected, or mathematically imprecise. 0.0 means E-level: no usable conjecture; incoherent, unrelated, essentially only fluff, or obviously false in simple cases."
    \item "Use intermediate values when appropriate. For example, a score of 0.6 may indicate an output between C and B, while a score of 0.9 may indicate a very strong output that is slightly below ideal."
\end{itemize}

\section{Pipeline Prompt Protocol}
\label{app:prompt-protocol}

The conjecture-generation pipeline used three prompts in sequence:
(1) theorem classification and theorem-specific object-map construction,
(2) theorem-specific redaction, and
(3) blind conjecture generation. This appendix summarizes the substantive
instructions that governed model behavior. We omit administrative instructions
concerning directory paths, filenames, Markdown formatting, identifier
bookkeeping, summary files, and behavior when no input files were present.

\subsection{Theorem Classification and Object-Map Construction}
\label{app:theorem-classification-prompt}

The first prompt instructed the model to inspect a paper, classify its
theorem-like environments, and identify its main proved theorems. The number of
qualifying results was not specified in advance, and the model was explicitly
instructed not to force a paper to contain a fixed number of main theorems.

A result qualified as a \emph{main proved theorem} only when it satisfied all
of the following criteria:
\begin{enumerate}
    \item it was presented as a contribution of the current paper;
    \item it had a proof in the current paper;
    \item it was not merely quoted from prior work;
    \item it was not a corollary, lemma, proposition, example, remark,
    definition, or background result;
    \item it played a central role in the paper's stated contribution.
\end{enumerate}
Evidence could include emphasis in the abstract or introduction, language such
as ``we prove,'' and the presence of a proof in the paper. The model separately
classified quoted prior theorems, supporting results, corollaries, and unproved
or deferred theorems. Ambiguous cases were to be marked as unclear rather than
treated as main proved theorems.

For each qualifying result, the model constructed a separate theorem-specific
object map. The object map contained:
\begin{itemize}
    \item the exact source theorem, its location, and its proof status;
    \item mathematical objects relevant to the result;
    \item required definitions and notation;
    \item assumptions and structural conditions;
    \item supporting lemmas, propositions, examples, constructions, quoted
    results, and background facts;
    \item an explicit or inferred proof-dependency chain;
    \item a concise proof-strategy summary;
    \item notes identifying material that should later be removed or
    generalized.
\end{itemize}
An object could appear in more than one map when it was relevant to multiple
theorems. At this stage, theorem statements and proof information were
permitted because the maps served as unredacted intermediate representations.
The model was instructed to preserve the source notation, use only the supplied
paper, avoid web search, and not invent theorem statements, labels, or proof
dependencies.

\subsection{Theorem-Specific Redaction}
\label{app:redaction-prompt}

The second prompt converted each object map independently into a redacted seed
for blind conjecture generation. Its core instruction was:
\emph{keep setup, remove answers}. The seed was intended to represent the local
mathematical setting surrounding one source theorem without revealing the
theorem itself.

The prompt permitted the seed to retain or non-conclusively rewrite:
\begin{itemize}
    \item the ambient mathematical setting;
    \item definitions and notation;
    \item fields, rings, groups, spaces, graphs, categories, probability
    spaces, and other domain objects;
    \item assumptions on parameters;
    \item relevant subsets, maps, operators, relations, functions, measures,
    metrics, bases, generators, and constructions;
    \item hypotheses and broad structural constraints;
    \item examples and auxiliary objects that did not reveal the conclusion;
    \item broad, setup-grounded directions along which conjectures might be
    formed.
\end{itemize}

The prompt required complete removal of:
\begin{itemize}
    \item the source theorem statement, conclusion, label, and any revealing
    title;
    \item statements describing what the theorem proves;
    \item the number or identity of the paper's main proved theorems;
    \item proof summaries, proof sketches, dependency chains, and exact
    logical paths leading to the theorem;
    \item result classifications and descriptions identifying a result as
    main, proved, central, or the paper's primary goal;
    \item theorem-concluding classifications, decompositions, equivalences,
    isomorphisms, bounds, rates, formulas, or existence claims;
    \item descriptions whose purpose was to explain an object's role in the
    theorem conclusion.
\end{itemize}

When theorem-dependent text also contained useful setup information, the model
was instructed to rewrite it as broader, non-conclusive context rather than
copy it. For example, a claim that an invariant equals a particular formula
could be reduced to the statement that the setup contains an invariant
associated with the objects under study. If information might reveal the
conclusion, the prompt instructed the model to remove or generalize it and to
prefer under-inclusion over leakage.

Each resulting seed was organized into the six context components used in the
ablation study:
\begin{enumerate}
    \item Ambient Setting;
    \item Core Definitions and Notation;
    \item Local Mathematical Objects;
    \item Structural Constraints and Conditions;
    \item Auxiliary Objects and Constructions;
    \item Potential Generalization Axes.
\end{enumerate}
The Potential Generalization Axes were to remain generic and grounded in the
redacted setup. Suggested directions included varying parameters, weakening or
strengthening assumptions, changing the ambient object or target class,
replacing a structure with an analogue, and examining small or exceptional
cases.

The redaction output also contained a forbidden-content section and a
redaction audit checking whether theorem statements, conclusions, proof
information, exact classifications, and result-count information had been
removed. Before conjecture generation, these two administrative sections were
manually removed, leaving only the six mathematical-context components as
candidate model input.

\subsection{Blind Conjecture Generation}
\label{app:generation-prompt}

The third prompt instructed the model to generate exactly one plausible formal
conjecture for each redacted seed. Under an ablation condition, the model
received only the context components retained by that condition. It was not
permitted to consult the original paper, \LaTeX{} source, theorem map,
unredacted object map, theorem statement, proof summary, dependency chain,
redaction report, or any external source.

For each seed, the model was instructed to choose the single most natural
conjecture suggested by the available local objects and constraints. The
conjecture was written in \LaTeX{} and could use definitions, notation,
objects, assumptions, examples, and structural constraints present in the
seed. The prompt permitted natural mathematical extrapolation, including a
classification, generalization, strengthening, weakening, analogue,
exceptional-case statement, structural claim, existence or uniqueness claim,
or dimension or bound conjecture. A newly proposed formula was permitted only
when it was derived from the redacted setup rather than copied from an
unredacted result.

The prompt stated that proximity to the hidden theorem was acceptable, and
even desirable, provided that the conjecture followed only from the redacted
seed. At the same time, the model was prohibited from claiming that the paper
proved the conjecture, claiming knowledge of the hidden theorem, copying an
exact conclusion from an unredacted source, or generating multiple candidate
conjectures for one seed. Information from one theorem-specific seed could not
be used for another unless it appeared independently in both.

For each seed, the requested response contained one precise formal statement,
the seed objects and assumptions used, a conjecture-type classification, a
plausibility rationale, a tentative proof direction, a counterexample-search
strategy, an assessment of possible proximity to a hidden source result, and
a blindness and leakage audit. The formal statement was treated as the
generated conjecture in the subsequent evaluation.

\section{Complete Ablation Results}
\label{sec:complete-ablation-results}

Table~\ref{tab:complete-ablation-results} reports the complete
leave-one-out and single-component results.

\begin{table*}[t]
\centering
\small
\setlength{\tabcolsep}{4pt}

\textbf{(a) Leave-one-out conditions}

\vspace{3pt}

\begin{tabular}{llrrrrrr}
\toprule
\textbf{Metric} &
\textbf{Removed component} &
\textbf{Full} &
\textbf{Ablated} &
\(\boldsymbol{\Delta}\) &
\textbf{95\% CI} &
\(\boldsymbol{d_z}\) &
\(\boldsymbol{p_{\mathrm{Holm}}}\) \\
\midrule
Quality
& Ambient setting
& .931 & .923 & -.008 & [-.028, .011] & -.108 & 1.000 \\
& Definitions and notation
& .931 & .931 & .000 & [-.022, .022] & .000 & 1.000 \\
& Local mathematical objects
& .931 & .923 & -.008 & [-.030, .013] & -.097 & 1.000 \\
& Structural constraints
& .931 & .910 & -.021 & [-.051, .008] & -.187 & 1.000 \\
& Auxiliary objects
& .931 & .938 & .007 & [-.013, .026] & .085 & 1.000 \\
& Generalization axes
& .931 & .936 & .005 & [-.012, .022] & .074 & 1.000 \\
\midrule
Alignment
& Ambient setting
& .607 & .582 & -.025 & [-.084, .035] & -.106 & 1.000 \\
& Definitions and notation
& .607 & .572 & -.034 & [-.110, .042] & -.116 & 1.000 \\
& Local mathematical objects
& .607 & .575 & -.031 & [-.105, .042] & -.108 & 1.000 \\
& Structural constraints
& .607 & .520 & -.087 & [-.154, -.020] & -.333 & .055 \\
& Auxiliary objects
& .607 & .579 & -.028 & [-.109, .053] & -.088 & 1.000 \\
& Generalization axes
& .607 & .641 & .034 & [-.038, .107] & .121 & 1.000 \\
\bottomrule
\end{tabular}

\vspace{12pt}

\textbf{(b) Single-component conditions}

\vspace{3pt}

\begin{tabular}{lrrr}
\toprule
\textbf{Component retained} &
\textbf{Quality} &
\textbf{Alignment} &
\textbf{Alignment gap} \\
\midrule
Ambient setting             & .885 & .348 & -.259 \\
Definitions and notation    & .889 & .397 & -.210 \\
Local mathematical objects  & .898 & .457 & -.149 \\
Structural constraints      & .851 & .348 & -.259 \\
Auxiliary objects           & .844 & .270 & -.336 \\
Generalization axes         & .820 & .266 & -.341 \\
\bottomrule
\end{tabular}

\caption{Complete ablation results. Panel (a) reports theorem-level
leave-one-out comparisons, and Panel (b) reports standalone performance
under the single-component conditions.}
\label{tab:complete-ablation-results}
\end{table*}

\section{Complete Domain-Level Results}
\label{sec:complete-domain-results}

Table~\ref{tab:complete-domain-results} reports the complete equal-weight paper-level domain summaries for both evaluation metrics, including all leave-one-out effects. The main-text summary in Table~\ref{tab:domain-alignment-summary} reports only the largest alignment loss and strongest standalone component within each domain. Each domain contains three papers, so these comparisons are exploratory.

\begin{table*}[t]
\centering
\small
\setlength{\tabcolsep}{5pt}

\textbf{(a) Theorem-conditioned conjecture quality}

\vspace{3pt}

\begin{tabular}{lrrrrrrr}
\toprule
\textbf{Domain} &
\textbf{Full} &
\textbf{Ambient} &
\textbf{Definitions} &
\textbf{Local objects} &
\textbf{Constraints} &
\textbf{Auxiliary} &
\textbf{Generalization} \\
\midrule
\texttt{math.AC} & .934 & -.030 & -.008 & -.018 & -.008 &  .010 & -.003 \\
\texttt{math.AG} & .881 &  .000 &  .029 &  .019 & -.037 &  .000 &  .016 \\
\texttt{math.CO} & .950 &  .006 & -.019 & -.006 & -.008 &  .017 &  .022 \\
\texttt{math.ST} & .986 & -.018 & -.008 & -.059 & -.014 & -.020 & -.014 \\
\bottomrule
\end{tabular}

\vspace{12pt}

\textbf{(b) Expected-theorem alignment}

\vspace{3pt}

\begin{tabular}{lrrrrrrr}
\toprule
\textbf{Domain} &
\textbf{Full} &
\textbf{Ambient} &
\textbf{Definitions} &
\textbf{Local objects} &
\textbf{Constraints} &
\textbf{Auxiliary} &
\textbf{Generalization} \\
\midrule
\texttt{math.AC} & .611 & -.037 & -.061 &  .004 & -.162 &  .011 & -.047 \\
\texttt{math.AG} & .571 & -.052 & -.053 & -.060 & -.121 & -.087 & -.046 \\
\texttt{math.CO} & .644 & -.003 & -.086 & -.042 & -.047 &  .019 &  .103 \\
\texttt{math.ST} & .664 &  .026 &  .073 & -.096 & -.011 & -.054 &  .162 \\
\bottomrule
\end{tabular}

\caption{Equal-weight paper-level results by domain. The full-context column
reports the domain baseline; the remaining columns report leave-one-out
differences from that baseline. Negative values indicate lower performance
after removing the component. Each domain contains three papers.}
\label{tab:complete-domain-results}
\end{table*}

\end{document}